\documentclass[10pt,twocolumn,letterpaper]{article}

\usepackage[pagenumbers]{cvpr} 

\usepackage{wrapfig}
\usepackage{multirow}
\usepackage[most]{tcolorbox}
\newtcolorbox{promptboxalt}[1]{colback=gray!10!white,colframe=black!80!white,title=#1}
\usepackage{amsthm}
\usepackage{xcolor}
\usepackage[framemethod=tikz]{mdframed}

\definecolor{obs-line}{HTML}{2E5EAA} 

\newtcolorbox[auto counter, number within=section]{observation}[1][]{
    enhanced,
    frame hidden,                    
    borderline west={3pt}{0pt}{obs-line}, 
    colback=obs-line!5!white,        
    coltitle=obs-line,               
    fonttitle=\bfseries,             
    title={Observation \thetcbcounter.},
    attach title to upper={\quad},   
    sharp corners,
    left=10pt, right=10pt, top=5pt, bottom=5pt,
    #1
}
\definecolor{cvprblue}{rgb}{0.21,0.49,0.74}
\usepackage[pagebackref,breaklinks,colorlinks,allcolors=cvprblue]{hyperref}
\usepackage{booktabs}
\usepackage{rotating}
\usepackage{adjustbox}

\def\paperID{*****} 
\def\confName{CVPR}
\def\confYear{2026}

\newcommand\countingtricks{\textsc{CountingTricks }}

\title{Counting to Four is still a Chore for VLMs}

\author{Duy Le Dinh Anh$^\dagger$, Patrick Amadeus Irawan$^\dagger$, Tuan Van Vo\\
MBZUAI\\
{\tt\small {\{duy.le, patrick.irawan, tuan.vo\}@mbzuai.ac.ae}}\\
{\small $^\dagger$Equal contribution.}
}

\begin{document}
\maketitle

\begin{abstract}
Vision--language models (VLMs) have achieved impressive performance on complex multimodal reasoning tasks, yet they still fail on simple grounding skills such as object counting. Existing evaluations mostly assess only final outputs, offering limited insight into where these failures arise inside the model. In this work, we present an empirical study of VLM counting behavior through both behavioral and mechanistic analysis. We introduce \countingtricks, a controlled evaluation suite of simple shape-based counting cases designed to expose vulnerabilities under different patchification layouts and adversarial prompting conditions. Using attention analysis and component-wise probing, we show that count-relevant visual evidence is strongest in the modality projection stage but degrades substantially in later language layers, where models become more susceptible to text priors. Motivated by this finding, we further evaluate \emph{Modality Attention Share} (MAS), a lightweight intervention that encourages a minimum budget of visual attention during answer generation. Our results suggest that counting failures in VLMs stem not only from visual perception limits, but also from the underuse of visual evidence during language-stage reasoning. Code and dataset will be released at \url{https://github.com/leduy99/-CVPRW26-Modality-Attention-Share}.
\end{abstract}
\section{Introduction}
\label{sec:intro}

Vision--Language Models (VLMs)~\cite{liu2023visual, instructblip, openai24} have been rapidly developing in recent times. VLMs integrate images into large language models (LLMs) and leverage the powerful reasoning abilities of these foundations~\cite{touvron2023llama2, zheng2023judging}, showcasing remarkable proficiency in tasks such as open-ended captioning, visual question answering, and instruction following. In particular, recent frontier models like GPT\cite{openai24} and Gemini~\cite{cbs+25} have regularly competed and stamped new SOTA performances in just months intervals.

\begin{figure}[t] 
    \centering
    \includegraphics[width=\linewidth]{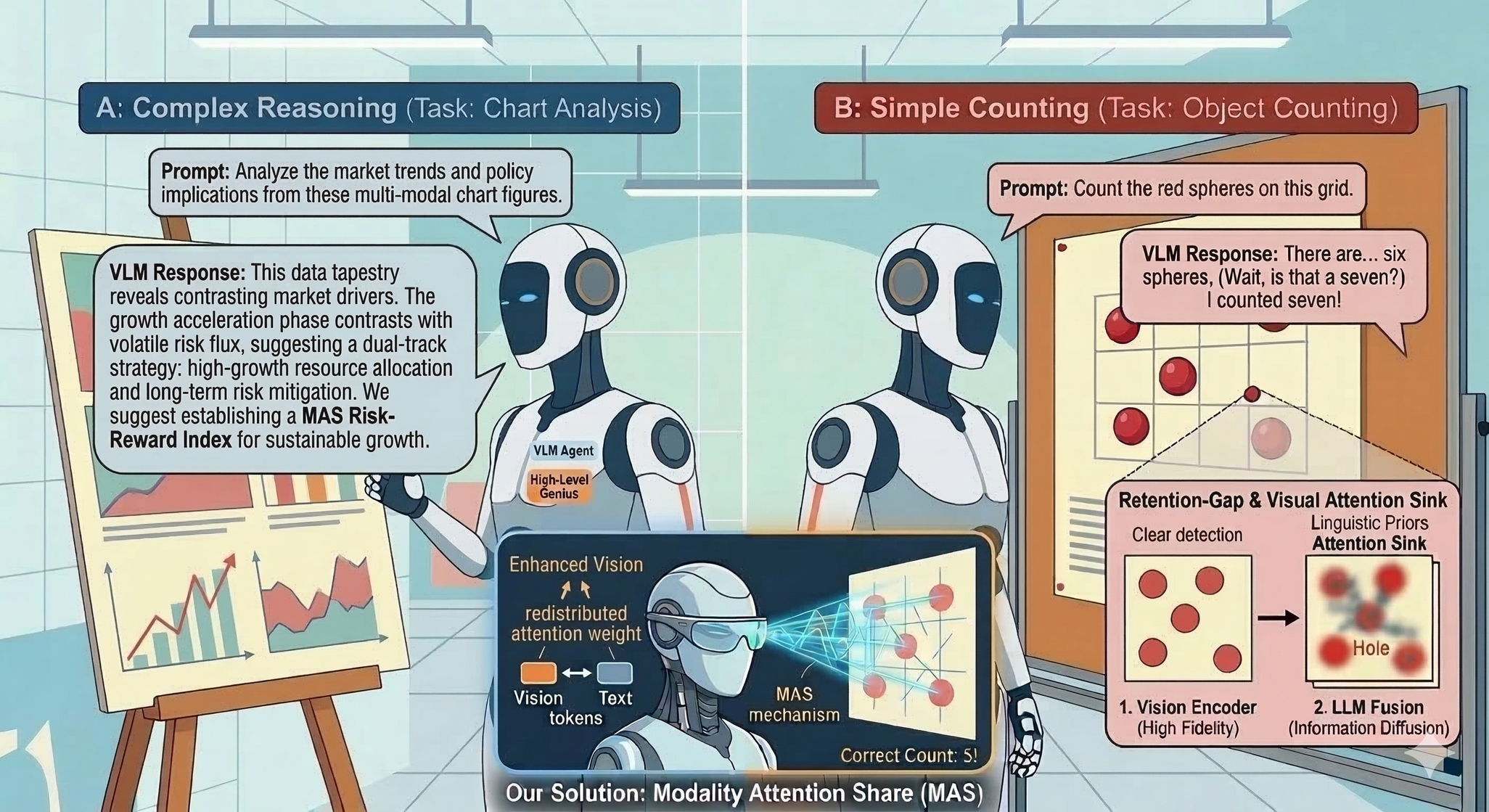} 
    \caption{
\textbf{Diagnosing the grounding gap in VLM counting.}
VLMs can answer complex multimodal questions, yet often fail at a minimal grounding skill like counting. The figure illustrates our diagnosis: counting evidence is present in early visual representations but becomes under-used as generation relies more on textual priors. We also test an attention-budget constraint (MAS) as a lightweight step toward improving grounding.
}

    \label{fig:main_teaser}
\end{figure}

Despite the advancements, a notable weakness still exists: these models continue to exhibit visual deficiencies on elementary problems, especially those that require spatial understanding and object awareness. One example is the object counting task, where contemporary VLMs often miscount even in plain settings, and this performance significantly worsens when they are presented with mild clutter or simple adversarial textual cues. Since it is interesting to see how textual context steers these inaccuracies, we raise a foundational question: {\em Do VLMs count based on what they are truly seeing or on their learned priors, and specifically in which component does this occur?}. In this work, we aim to answer such a question and point out the exact case and source of this shortcoming within VLMs.

To support our study, we curate the \textbf{\countingtricks} evaluation suite, specifically designed to test basic object counting ability across multiple patchification settings, including object layout and object shape, thereby verifying the known counting deficiency in SOTA open-weight and commercial VLMs. To advance this analysis to a deeper interpretability level, we introduce a probing framework that attaches lightweight detection heads to the vision encoder, modality projector, and backbone LLM to pinpoint the exact bottleneck component; this layer-wise diagnosis reveals that errors stem from a failure to retain visual evidence against linguistic dominance, not a lack of initial ``seeing.'' Finally, to alleviate this visual information imbalance, we propose \textbf{Modality Attention Share (MAS)}, a technique for attention redistribution between modalities via regularization, which yields consistent accuracy gains by successfully ``forcing'' the model to re-attend to the visual context during complex object counting tasks.

\begin{enumerate}
    \item We introduce \countingtricks, an object-counting evaluation suite containing 18k cases across 32 different patchification settings, systematically covering variants in object size, shape, and adjacency.
    \item We conduct a surface and deep-level interpretability study to pinpoint inaccuracies at the token level, as well as to pinpoint the core component bottleneck that causes counting ability deficiency across open-weight models.
    \item We propose Modality Attention Share (MAS), which redistributes attention across modalities. Our findings suggest that we are able to obtain up to a 1\% consistent accuracy gain by allowing the VLM to re-attend more to visual tokens.
\end{enumerate}

\section{Related Work} \label{sec:related}

\noindent\textbf{Vision-Language Models.} Recent Vision-Language Models (VLMs) have achieved remarkable general-purpose reasoning by integrating powerful LLMs with pre-trained vision encoders~\cite{liu2023visual, instructblip, openai24}. However, despite this success, fine-grained spatial tasks like object counting remain a persistent failure mode. While specialized architectures like FSC-147~\cite{ranjan2021learning} handle counting robustly, general-purpose VLMs often struggle to replicate this precision. Recent innovations attempt to address this by rethinking tokenization—such as dynamic resolution in Qwen2.5-VL~\cite{qwen2025vl} or sub-image partitioning in LVLM-COUNT~\cite{lvlmcount2025}. However, studies on token compression~\cite{wei2024treat, zuo2024dynamic} suggest that VLMs can often perform tasks even with aggressively pruned visual features. This implies a structural tendency to rely on semantic text priors over dense spatial evidence, a hypothesis our work investigates by tracing information loss across model components.
\vspace{2mm}

\noindent\textbf{Diagnostic Benchmarks for Vision-Language Models.} To rigorously assess these limitations, the field has moved from general VQA benchmarks (e.g., TextVQA, GQA) to targeted diagnostic suites. A growing body of work reveals that VLMs stumble on visually trivial skills~\cite{tong2025vlms}. For instance, \emph{VLMs Can't See the Obvious}~\cite{tong2025vlms} exposes systematic errors on basic attributes, while specific counting benchmarks like \emph{PairTally}~\cite{ma2025pairtally} and \emph{VLMCount Bench}~\cite{guo2025vlmcount} highlight failures in distinguishing fine-grained pairs. Crucially, the most effective diagnostics focus on the tension between visual evidence and linguistic priors. Studies such as \emph{Blind Faith in Text}~\cite{jeong2025blind} and \emph{When Language Overrules}~\cite{kim2025language} demonstrate that modest adversarial perturbations or conflicting instructions can cause models to ignore visual data entirely. These benchmarks highlight a critical vulnerability: models often bypass perceptual reasoning in favor of probabilistic language generation. We advance this line of inquiry with \textbf{\countingtricks}, which goes beyond surface-level accuracy to probe internal failure modes linked to patchification layouts and object adjacency, serving as a testbed for our component-level analysis.

\vspace{2mm}
\noindent\textbf{Interpretability in Vision-Language Models.} While benchmarks identify \emph{that} models fail, our work seeks to understand \emph{why} and \emph{where}. A critical line of inquiry focuses on "Modality Imbalance," where textual priors dominate visual evidence~\cite{zhang2024words, wan2024two}. Probing studies such as \emph{What’s in the Image?}~\cite{zhang2024whats} report that spatial details are often diluted across fusion layers, creating a "visual attention sink." To mitigate this, training-free methods like VAR~\cite{var2025} propose attention redistribution. We build upon these insights but take a more rigorous approach: instead of just observing the imbalance, we operationalize the \textbf{Modality Attention Share (MAS)} as a differentiable training objective. This allows us to actively correct the imbalance during fine-tuning, directly reducing the root cause of the visual information loss we identify in our probing experiments.

\section{CountingTricks Evaluation Suite}\label{sec:bench}

In this section, we detail the construction of the \countingtricks evaluation suite. We first describe the programmatic generation of the visual suite, designed to systematically cover VLMs behavior when presented by different counting case layouts. Then, we define the evaluation metrics that allow us to evaluate this counting ability degradation on coarse-level (accuracy) until more fine-grained (attn-IoU).


\subsection{Image and Prompt Construction} \label{sec:image-prompts}

\begin{figure*}[t]
    \centering
    \includegraphics[width=0.99\linewidth]{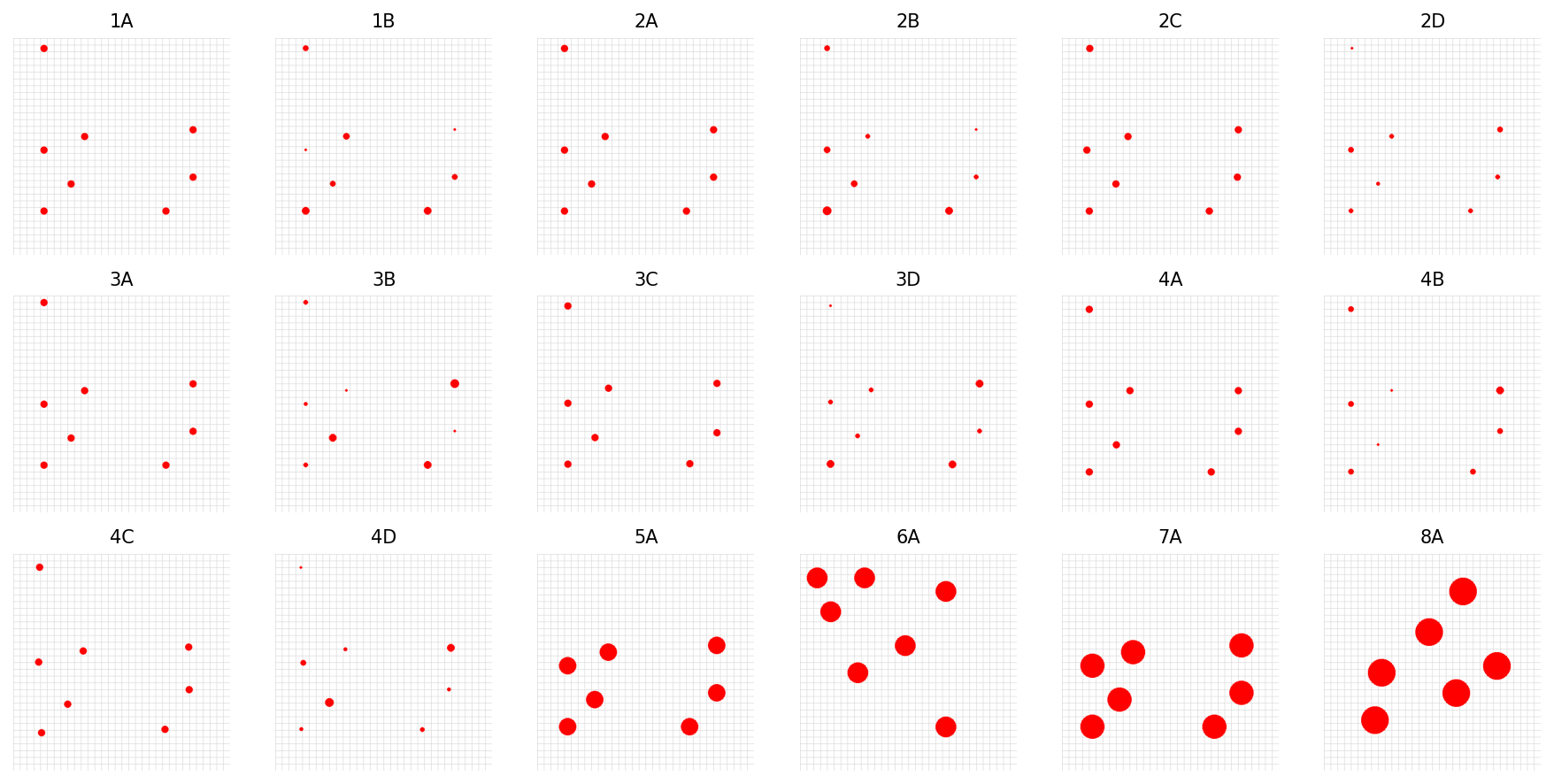}\\
    \caption{\textbf{The Toy-32 Visual Suite.} We craft multiple patchification cases to probe perception failures under different geometric alignments and isolate Vision Encoder blind spots. Object Counting Benchmark Data: 1A–B (fixed/varied sizes), 2A–D (vertical grid alignment), 3A–D (horizontal grid alignment), 4A–D (grid intersection alignment), 5A–8A (circle diameters 2.5--4 patch sizes).}
    \label{fig:toy32}
\end{figure*}

The \countingtricks benchmark evaluates basic VLM perceptual ability, similar to \citet{vlmblind}, but focused on object counting. We generate all samples programmatically to balance object count, shape, color, and placement across patchification cases. Shapes include squares, triangles, and circles, with object counts $N \in [3, 12]$. \countingtricks varies the object's spatial relationship to the Vision Encoder patch grid (numerical prefix) and adds size/positional complexity (alphabetical suffix), yielding 32 cases (Fig.~\ref{fig:toy32}).

\paragraph{Numerical Prefixes (Positioning and Patchification Setting).} The numbers \textbf{1} through \textbf{4} define the object's center alignment relative to the patch grid, testing feature fragmentation across patches.
\begin{itemize}
    \item \textbf{(1) Cell-centered}: Object is centered entirely within a patch (Best-case alignment).
    \item \textbf{(2) Vertical Grid Line Alignment (Grid-Alignment)}: Object is centered on the vertical line between two adjacent patches.
    \item \textbf{(3) Horizontal Grid Line Alignment}: Object is centered on the horizontal line between two adjacent patches.
    \item \textbf{(4) Center Intersection (Intersection)}: Object is centered exactly at the intersection of four patches (Worst-case quantization).
\end{itemize}

\paragraph{Alphabetical Suffixes (Size and Translation).} The letters \textbf{A}, \textbf{B}, \textbf{C}, and \textbf{D} define variations in size heterogeneity and positional noise applied to the core placement types (1--4).
\begin{itemize}
    \item \textbf{(A) Fixed Size}: The base case; all objects have the same, small fixed diameter.
    \item \textbf{(B) Varying Size}: Objects have heterogeneous diameters, randomly resized down to $20\%$ of the base size.
    \item \textbf{(C \& D) In-Position Translation}: Minor random positional jitter around the defined placement center.
    \begin{itemize}
        \item \textbf{C}: Fixed Size with Translation.
        \item \textbf{D}: Varied Size with Translation.
    \end{itemize}
\end{itemize}

\paragraph{Special Size Variation and Adjacency Test Cases (5--15)}
These cases test failure modes related to scale and object density.
\begin{itemize}
    \item \textbf{5--8 (Varying Size/Dilation)}: Cell-centered placement (Type 1) with larger circle diameters (2.5--4$\times$ patch size; e.g., 5A--8A) to test scale robustness.
    \item \textbf{9--15 (Density/Adjacency Analysis)}: Objects are clustered with high adjacency (minimal gaps) to challenge instance separation.
\end{itemize}

\paragraph{Prompt} To probe text-prior dominance, we pair images with adversarial prompts combining \textless object\textgreater, \textless false\_count\textgreater\ (a distractor number), and \textless color\textgreater. The corresponding prompt templates are given below:

\begin{promptboxalt}{Standard Counting (Control) $P_{std}$}
How many \textless object\textgreater~ are there in the image?

Respond concisely with shape counts using the following format:
``\textless object\textgreater: \{number\}''.
For example: ``\textless object\textgreater: 7'' (example only).

[image: \textless quantity\textgreater~ of \textless object\textgreater]
\end{promptboxalt}

\noindent Here, [image: ...] denotes the actual input image. In the `Digit-in-Conflict' setting, \textless false\_count\textgreater~ is set to $N \pm 1$ or $N \pm 2$. A visually grounded model should reject the premise; a language-biased model often shifts the count toward the suggestion.

\subsection{Evaluation Models \& Metrics} \label{sec:models-metrics}
\textbf{Models.} We evaluate 10 SOTA open-weight VLMs, including both legacy and newer baselines. We specifically target the $\mathbf{3B}$--$\mathbf{11B}$ parameter range, which represents the most widely deployed class of models for edge and local inference. All model choices and their parameters can be observed in Table~\ref{tab:models}.

\begin{table}[!ht]
    \centering
    \caption{\textbf{Key Details of the Evaluated Vision-Language Models.} We focus on recent efficient open-weights models (3B--11B parameters) to test the accessibility of counting capabilities.}
    \resizebox{0.99\linewidth}{!}{
    \begin{tabular}{lcccc}
        \toprule
        \textbf{Model} & \textbf{Year} & \textbf{Backbone} & \textbf{\# Params} \\
        \midrule
        LLaVA-1.5-7B & 2023 & Vicuna-7B & 7B \\
        LLaVA-1.6-Vicuna-7B & 2024 & Vicuna-7B & 7B \\
        InternVL3-8B & 2025 & InternViT & 8B \\
        Llama-3.2-11B & 2024 & Llama-3.2 & 11B \\
        Gemma-3-4B & 2025 & Gemma-3 & 4B \\
        Qwen2.5-VL-3B & 2025 & Qwen2.5 & 3B \\
        Phi-4-Multimodal & 2025 & Phi-4 & 5B \\
        LLaVA-OneVision & 2024 & Qwen2 & 7B \\
        Ovis-8B & 2024 & Llama-3 & 8B \\
        Qwen2.5-VL-7B & 2025 & Qwen2.5 & 7B \\
        \bottomrule
    \end{tabular}
    }
    \label{tab:models}
\end{table}

\paragraph{Evaluation Metrics.}
For each test sample in our benchmark, we use diagnostic metrics that trace the information flow. Let a single test sample be denoted as $q := (p, x, y)$, where $p$ is the input prompt, $x$ is the input image, and $y \in \mathbb{N}_+$ is the ground-truth count.

\begin{enumerate}
    \item \textbf{Accuracy.} Accuracy evaluates whether the model's final prediction matches the actual count. Specifically, accuracy checks whether the ground-truth number is present in the predicted output string. For example: if the actual count is $\mathbf{5}$ (Ground-Truth) and the model predicts ``There are five red apples,'' the accuracy is $1$. If the model predicts ``There are four red apples,'' the accuracy is $0$. To aid in the conversion process, we provide a lookup table for numbers from $\mathbf{1}$ until $\mathbf{30}$. This table is used to directly convert word-based numbers (e.g., ``one,'' ``fifteen,'' ``thirty'') into their numerics (e.g., $1$, $15$, $30$).
    
    \item \textbf{Attention Intersection over Union (Attn-IoU).} To measure the physical grounding of the model's attention, we back-project attention maps from component $C \in \{\mathrm{VE}, \mathrm{LLM}\}$ to the image grid. We compute the Intersection over Union (IoU) with the ground truth object mask $\mathrm{GT}$:
    {\small
    $$
    \mathrm{Attn\text{-}IoU}(C, {\cal Q}) := |{\cal Q}|^{-1} \sum_{q \in {\cal Q}} \frac{|\,\mathrm{AttnMask}_q(C) \cap \mathrm{GT}_q\,|}{|\,\mathrm{AttnMask}_q(C) \cup \mathrm{GT}_q\,|}.
    $$}
    Here, $\mathrm{AttnMask}_q(C)$ is the binarized top-$k\%$ attention map derived from component $C$. This metric quantifies how well the model "looks" at the correct objects.

    \item \textbf{Average Precision (AP).} Distinct from the generative metrics above, we utilize Average Precision to assess the quality of intermediate feature representations during our YOLO probing experiments. We adopt the standard object detection definition, specifically reporting \textbf{AP@50} (AP at an IoU threshold of 0.5). For a set of detections, AP is defined as the area under the Precision-Recall curve:
    $$
    \mathrm{AP} = \sum_{n} (R_n - R_{n-1}) P_n
    $$
    where $P_n$ and $R_n$ correspond to the precision and recall at the $n$-th threshold. In our diagnostic framework, a high AP indicates that the features at a specific model tap (e.g., Projector or LLM layers) retain sufficient spatial geometry to support object localization, independent of the model's final textual output.
\end{enumerate}

\section{Evaluation Results}
\label{sec:results}

\subsection{Overall Results}
\label{sec:baseline_analysis}
\paragraph{Baseline Performance Hierarchy.}
We first assess the capability of 10 VLMs on the \countingtricks{} benchmark. Table~\ref{tab:main_results} and Table~\ref{tab:main_results_details} (Appendix) present the accuracy across all 32 controlled regimes. The results indicate a clear generational shift in visual reliability. While legacy models such as LLaVA-1.5-7B perform poorly with an average accuracy of $11.82\%$, recent architectures incorporating advanced visual tokenizers, such as Qwen2.5-VL-7B, achieve a highest average of $50.52\%$. Notably, model scale is not the primary determinant of success; the efficient 3B parameter variant of Qwen2.5-VL ($36.01\%$) significantly outperforms the larger Llama-3.2-11B ($24.00\%$). This suggests that architectural decisions regarding resolution and spatial embedding preservation are more critical for counting tasks than parameter count alone.

\begin{table}[!ht]
    \centering
    \caption{\textbf{Baseline Evaluation on Counting-Tricks.} Models are ranked by average accuracy. Despite improvements in recent architectures (e.g., Qwen2.5-VL), most models struggle to exceed 50\% accuracy on this diagnostic suite. Each test case is evaluated using 1000 samples, evenly distributed across the respective count range.}
    \resizebox{0.85\linewidth}{!}{ 
    \begin{tabular}{lc}
        \toprule
        \textbf{Model} & \textbf{Avg Accuracy (\%)} \\
        \midrule
        LLaVA-1.5-7B & 11.82\% \\
        LLaVA-1.6-Vicuna-7B & 14.20\% \\
        InternVL3-8B & 21.72\% \\
        Llama-3.2-11B & 24.00\% \\
        Gemma-3-4B & 27.37\% \\
        Qwen2.5-VL-3B & 36.01\% \\
        Phi-4-Multimodal & 39.83\% \\
        LLaVA-OneVision-Qwen2-7B & 44.44\% \\
        Ovis-8B & 47.90\% \\
        \textbf{Qwen2.5-VL-7B} & \textbf{50.52\%} \\
        \bottomrule
    \end{tabular}
    }
    \label{tab:main_results}
\end{table}

\paragraph{Layout Sensitivity and Patchification.} Decomposing the performance by geometric configuration reveals systematic vulnerabilities linked to the vision encoder's patchification mechanism. We observe two distinct trends. First, regarding \textbf{Dilation Robustness}, performance consistently improves as object size increases relative to the patch grid. Comparing the baseline small objects (Case 1A, 39.08\% avg) to dilated variants (Case 5A, 52.98\% avg), models exhibit substantial gains (e.g., Qwen2.5-VL-7B improves from $56.3\%$ to $73.3\%$). This confirms that "patchification noise" is a primary failure mode for small objects. Conversely, we identify a severe \textbf{Adjacency Collapse} in high-density regimes. In Cases 9--15, where objects touch or closely abut, accuracy collapses across all models (e.g., Case 9B yields consistently low scores, with InternVL3-8B dropping to $8.9\%$). This indicates that current visual encoders struggle to resolve separate instances when the inter-object gap approaches the Nyquist limit of the patch grid.

\textcolor{blue}{\textbf{Observation 1.}} \textit{Our results reveal that current vision-language models fall short of reliable counting, exhibiting a marked sensitivity to count complexity. This observation suggests that when pixel-level evidence becomes ambiguous or fine-grained, models often abandon grounded perception to rely on linguistic priors.}

\paragraph{Correlation Analysis and Number Avoidance.}
We analyze the relationship between ground truth counts and model accuracy, finding a strong negative correlation (average $r \approx -0.78$) across all patchification regimes. This trend suggests that models do not accumulate spatial evidence linearly; instead, performance degradation accelerates as the count increases. As detailed in Figure \textcolor{blue}{8} (Appendix), this behavior manifests as ``Number Avoidance''—a distributional bias where models favor smaller or more frequent numbers found in their instruction-tuning data while systematically ignoring others. For instance, LLaVA-1.5-7B exhibits a complete collapse in capability for specific numerals, achieving $0.0\%$ accuracy for counts 7, 8, 9, and 11. Conversely, while the state-of-the-art Qwen2.5-VL-7B maintains robust accuracy for lower counts (e.g., $99.3\%$ for count 2) , it curiously fails completely at count 11 ($0.0\%$), despite recovering performance at count 12 ($20.1\%$). This implies the error is not purely a function of visual density, but of linguistic frequency. This brings us to a novel insight:

\textcolor{blue}{\textbf{Observation 2.}} \textit{The ``Number Avoidance'' phenomenon indicates that counting errors are not merely random noise but are systematically linked to linguistic priors. Models exhibit selective blindness toward specific integers (e.g., prime numbers like 7 or 11), suggesting that when visual evidence is ambiguous, the generation process defaults to high-probability tokens from the language model rather than grounding in the pixel data.}

\subsection{Attention Interpretability and Spatial Signal Analysis: Evidence of Text-Prior Dominance}
\label{sec:probing_analysis}

\begin{figure*}[t]
    \centering
    \begin{subfigure}[b]{0.48\textwidth}
        \includegraphics[width=\linewidth, height=5.5cm]{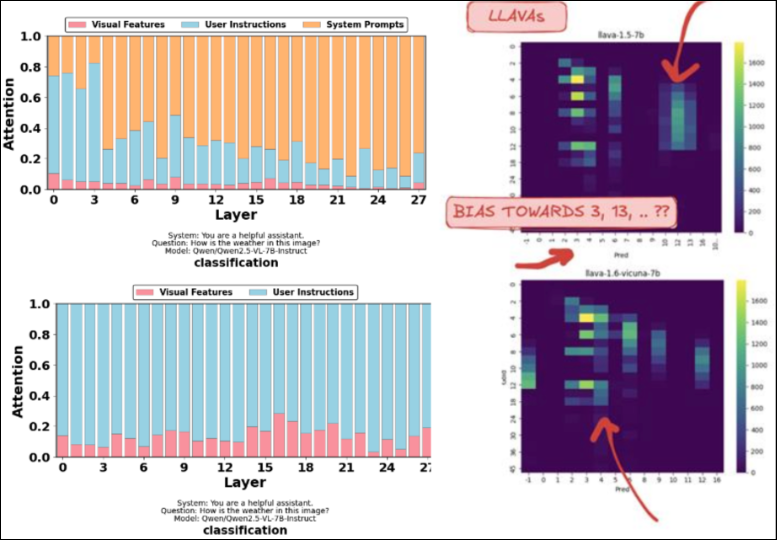}
        \caption{\textbf{Modality Attention Share.} Despite the task being purely visual, attention mass is overwhelmingly allocated to text tokens (blue bars), leaving only $\sim 10.7\%$ for visual tokens (red bars).}
        \label{fig:attn_share}
    \end{subfigure}
    \hfill
    \begin{subfigure}[b]{0.48\textwidth}
        \includegraphics[width=\linewidth, height=5.5cm]{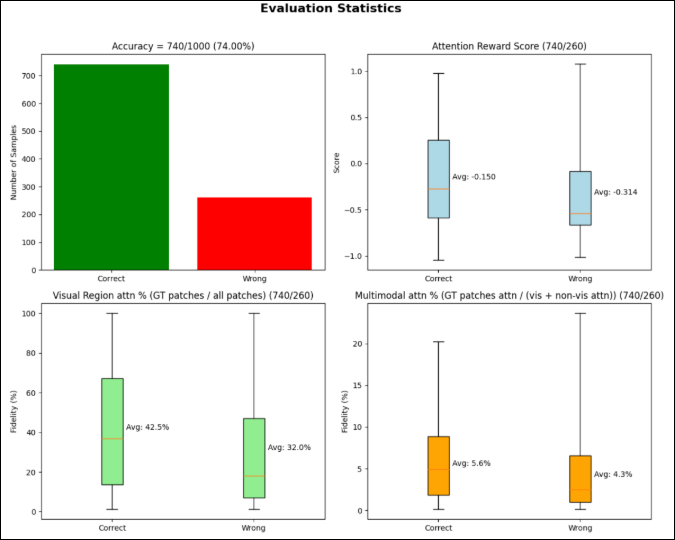}
        \caption{\textbf{Grounding Statistics.} Even for correct predictions (green), the Visual Region Attention \% is low ($\sim 42.5\%$) and the Attention Reward Score is negative ($-0.15$), indicating weak pixel commitment.}
        \label{fig:grounding_stats}
    \end{subfigure}
    \caption{\textbf{Evidence of Text-Prior Dominance.} Our probing analysis reveals that models suffer from a ``visual attention sink,'' where computation drifts away from image tokens toward system prompts and instructions.}
    \label{fig:attention_analysis}
\end{figure*}

To elucidate the internal mechanisms of counting failures, we analyze the internal feature maps using our probing framework. We visualize the attention distribution and grounding metrics in Figure~\ref{fig:attention_analysis}.

\paragraph{Modality Imbalance and Attention Sinks.}
Our component importance analysis reveals a stark imbalance in resource allocation. As shown in Figure~\ref{fig:attn_share}, the \emph{Modality Attention Share} in the LLM layers is heavily skewed toward text. On average, models allocate approximately $89.3\%$ of their attention budget to system prompts and instructions, leaving only $\sim 10.7\%$ for visual tokens. This ``visual attention sink'' implies that the generation process is driven largely by linguistic priors rather than pixel-level evidence. Attempts to steer this behavior with prompts (e.g., ``Please look at the image'') proved unreliable, often failing to shift the attention mass significantly.


\paragraph{The Illusion of Grounding.}
Finally, our Saliency-IoU analysis (Figure~\ref{fig:grounding_stats}) reveals that correct answers are often ungrounded. Even when models output the \emph{correct} number, their attention maps often fail to align with the ground truth objects. We observe a negative mean Attention Reward Score ($-0.15$) and a low median visual region attention ($\sim 42.5\%$). This implies that correct answers often emerge from probabilistic guessing rather than precise spatial accounting.

\textcolor{blue}{\textbf{Observation 3.}} \textit{Correct counts in VLMs are often ungrounded. The lack of attention overlap with ground truth objects suggests that models rely on "linguistic recall" from priors rather than accumulating evidence from pixels, rendering them fragile to adversarial text cues.}

\subsection{Where does visual evidence fade? Early vs Fused}
\label{sec:early_vs_fused}

To pinpoint exactly where the counting signal degrades, we compare the localization performance of our standardized YOLO probes attached to the Encoder, Projector, and LLM taps. Crucially, to ensure a fair comparison, we employ an identical lightweight probing architecture across all stages: a $1\times1$ bottleneck ($C_{\text{in}}\times 512 + \text{GroupNorm} + \text{SiLU}$) followed by a shared YOLO head.

As detailed in Table~\ref{tab:probe_params}, the total trainable parameters for each probe remain consistently low ($\sim 1.7$M to $2.2$M). The parameter delta between taps (e.g., $\sim 0.5$M due to input channel differences) is negligible compared to the frozen VLM backbone (billions of parameters), ensuring that any performance difference reflects the quality of the retained representation, not the probe's learning capacity.

\begin{table}[h]
\centering
\caption{\textbf{Probe Architecture Parameters.} We maintain negligible capacity differences between taps to ensure a fair apple-to-apples comparison of feature utility.}
\label{tab:probe_params}
\resizebox{0.9\linewidth}{!}{
\begin{tabular}{lccc}
\toprule
\textbf{Tap} & \textbf{Bottleneck} & \textbf{YOLO Head} & \textbf{Total Trainable} \\
\midrule
Encoder (1024ch) & $\sim 0.52$M & $\sim 1.19$M & $\sim \mathbf{1.71}$M \\
Projector (2048ch) & $\sim 1.05$M & $\sim 1.19$M & $\sim \mathbf{2.24}$M \\
LLM (2048ch) & $\sim 1.05$M & $\sim 1.19$M & $\sim \mathbf{2.24}$M \\
\bottomrule
\end{tabular}
}
\end{table}

\begin{figure}[t]
    \centering
    \includegraphics[width=\linewidth]{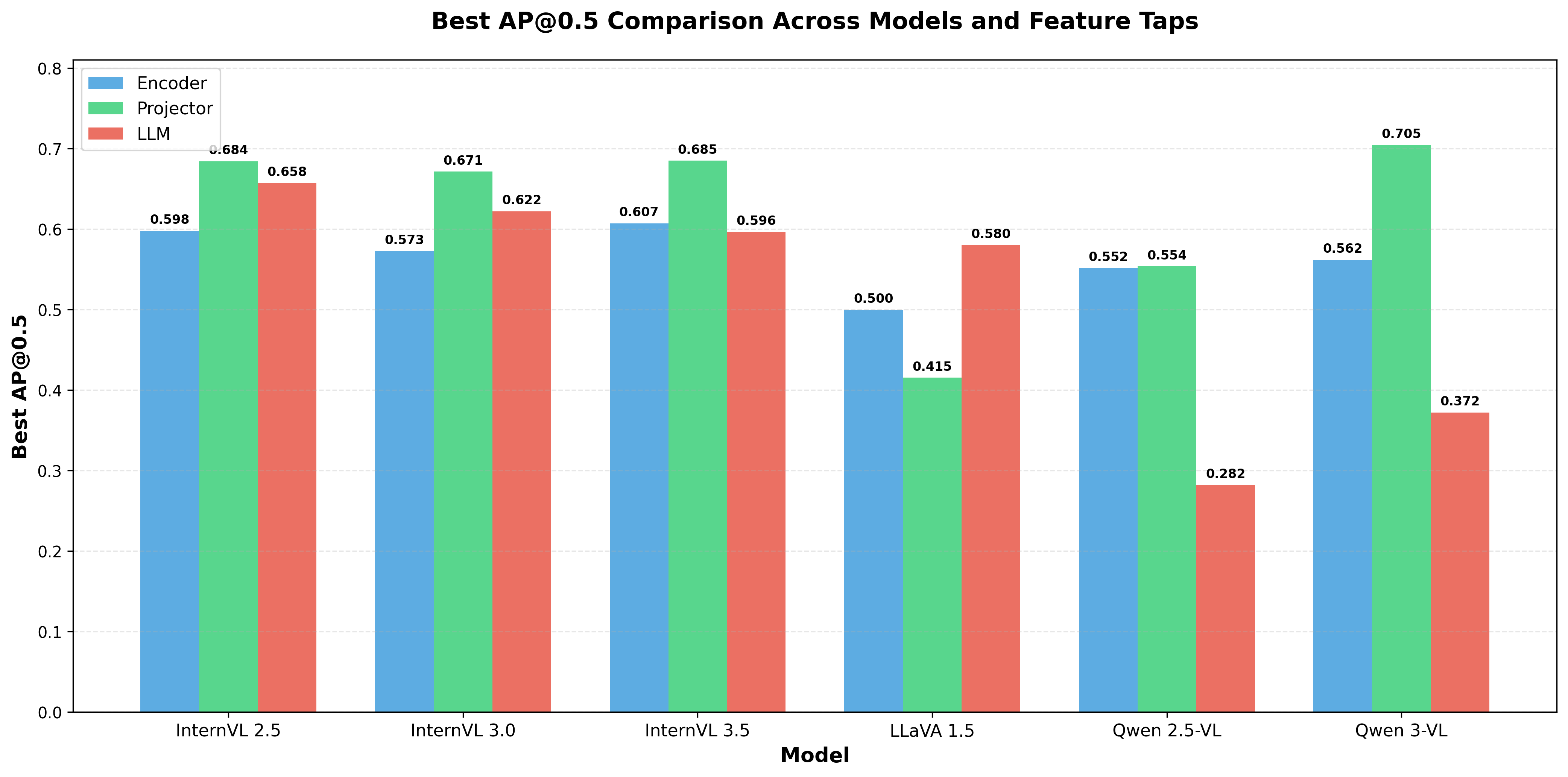}
    \caption{\textbf{Spatial evidence is strong in the vision stack but fades in the language layers.} Detection performance (AP@0.5) peaks at the projector tap (green) and drops at the LLM tap (red) across modern architectures, quantifying the loss of spatial information.}
    \label{fig:best_ap}
\end{figure}

\begin{figure*}[t]
    \centering
    \begin{subfigure}[b]{0.32\textwidth}
        \centering
        \includegraphics[width=\linewidth]{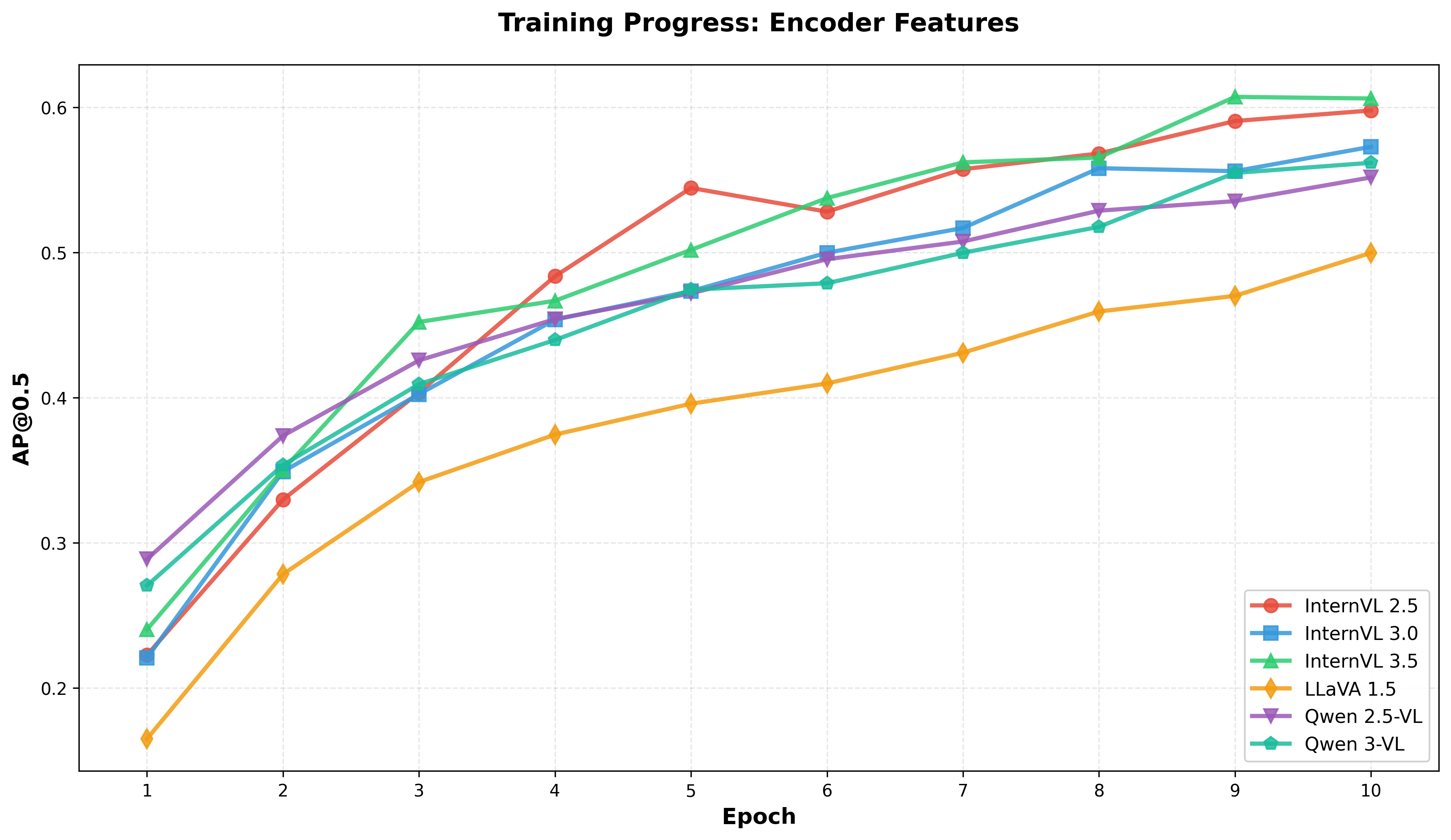}
        \caption{\textbf{Encoder}}
    \end{subfigure}
    \hfill
    \begin{subfigure}[b]{0.32\textwidth}
        \centering
        \includegraphics[width=\linewidth]{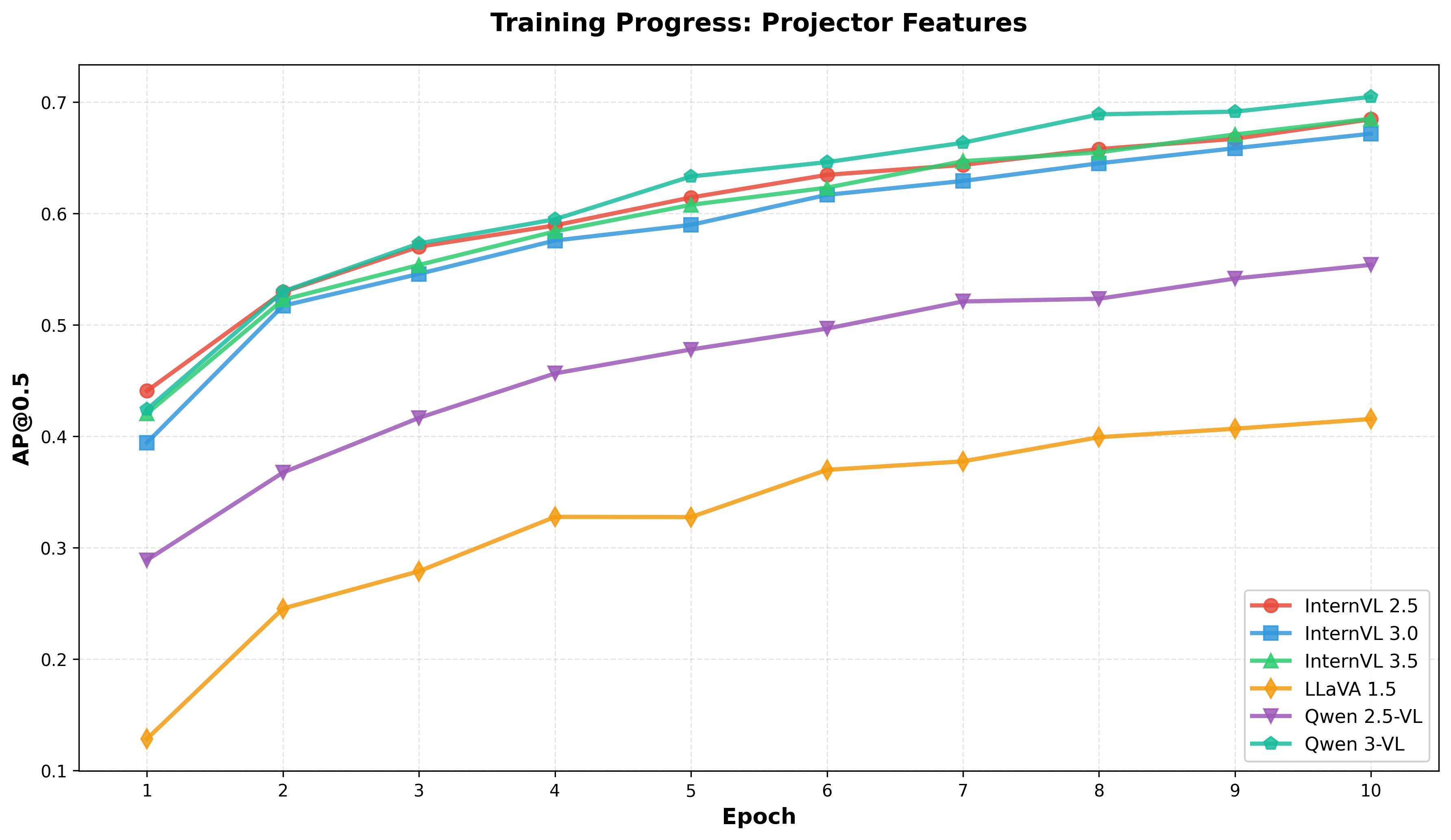}
        \caption{\textbf{Projector}}
    \end{subfigure}
    \hfill
    \begin{subfigure}[b]{0.32\textwidth}
        \centering
        \includegraphics[width=\linewidth]{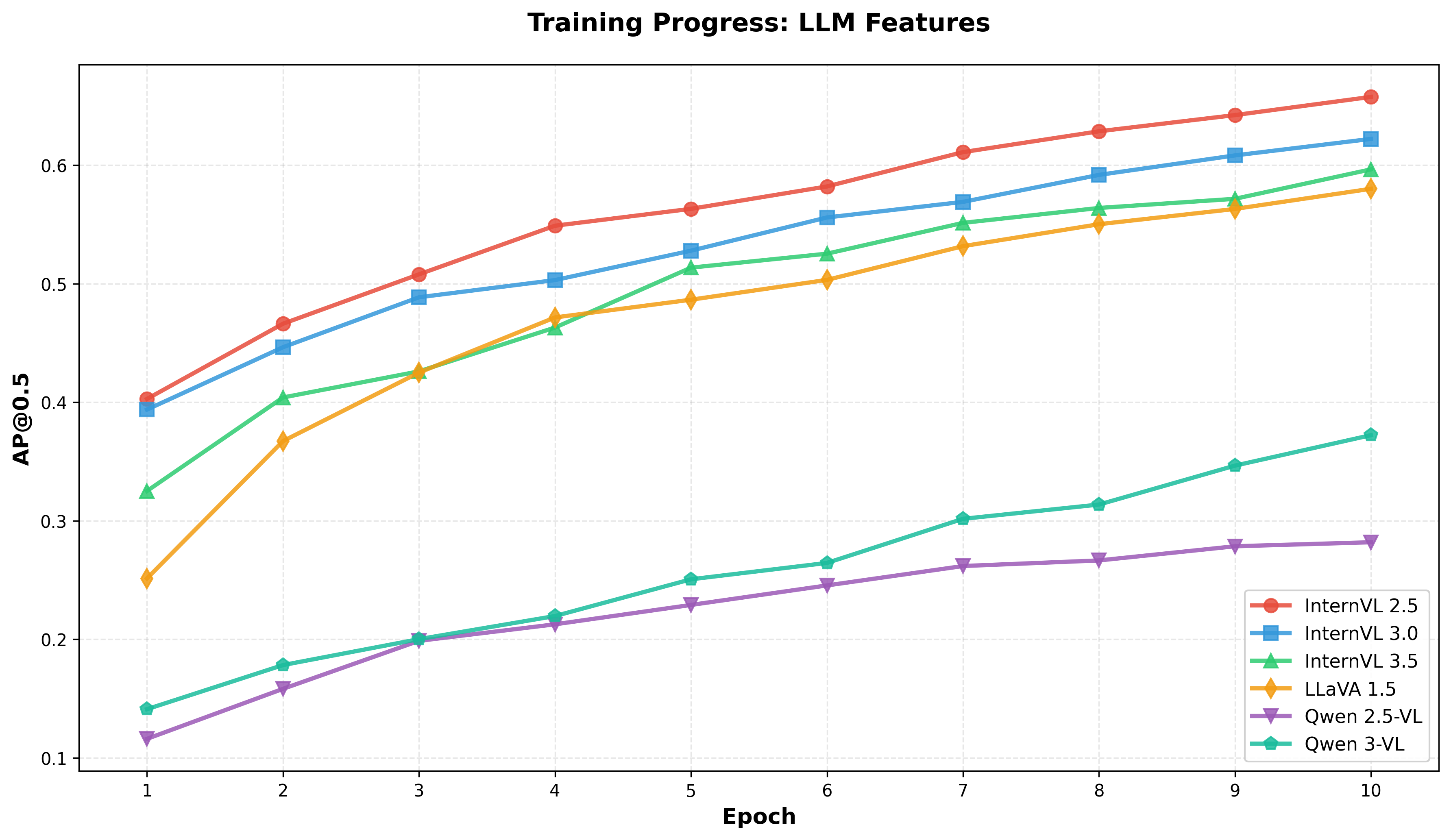}
        \caption{\textbf{LLM}}
    \end{subfigure}
    \caption{\textbf{Training Dynamics (YOLO Probing Head).} The Projector probe (center) reaches higher AP sooner, while the Encoder plateaus lower and LLM curves exhibit higher volatility. Since the schedule and seed are identical across taps, these differences reflect purely \emph{tap utility} rather than optimization quirks.}
    \label{fig:training_dynamics}
\end{figure*}

\noindent \textbf{Convergence Dynamics.}
We further analyze the learning trajectory of the probes to understand signal quality. As shown in Figure~\ref{fig:training_dynamics}, the Projector probe converges significantly faster and reaches a higher AP asymptote compared to the others. The LLM probe (right) exhibits slower convergence and higher volatility, suggesting that the spatial features at this stage are not only weaker but also structurally noisier.

\noindent \textbf{Quantifying signal fade from vision to language.} Despite an identical probing setup, we observe a consistent drop in detection performance from the projector to the LLM across multiple architectures (Figure~\ref{fig:best_ap}). Average Precision (AP) peaks at the projector tap but degrades at the LLM tap. For example, in Qwen2.5-VL, AP falls from $\mathbf{0.554}$ (projector) to $\mathbf{0.282}$ (LLM); in Qwen3-VL, from $\mathbf{0.705}$ to $\mathbf{0.372}$. These results indicate that while the visual backbone encodes object-level spatial information well, this signal is diluted during fusion and language reasoning.

\textcolor{blue}{\textbf{Observation 4.}} \textit{Spatial evidence is strong in the vision stack but fades in the language layers. The visual backbone encodes objects well (high AP at the projector), yet this spatial identity is largely lost during LLM reasoning (low AP at the LLM), indicating a failure of integration rather than perception.}


\paragraph{Qualitative Analysis of Visual Information Fading}

We conduct an analysis to visualize and understand the fading of visual information at the attention interpretability level as the signal passes from the image input through the different VLM components.

\begin{figure}[t]
    \centering
    \includegraphics[width=0.95\linewidth]{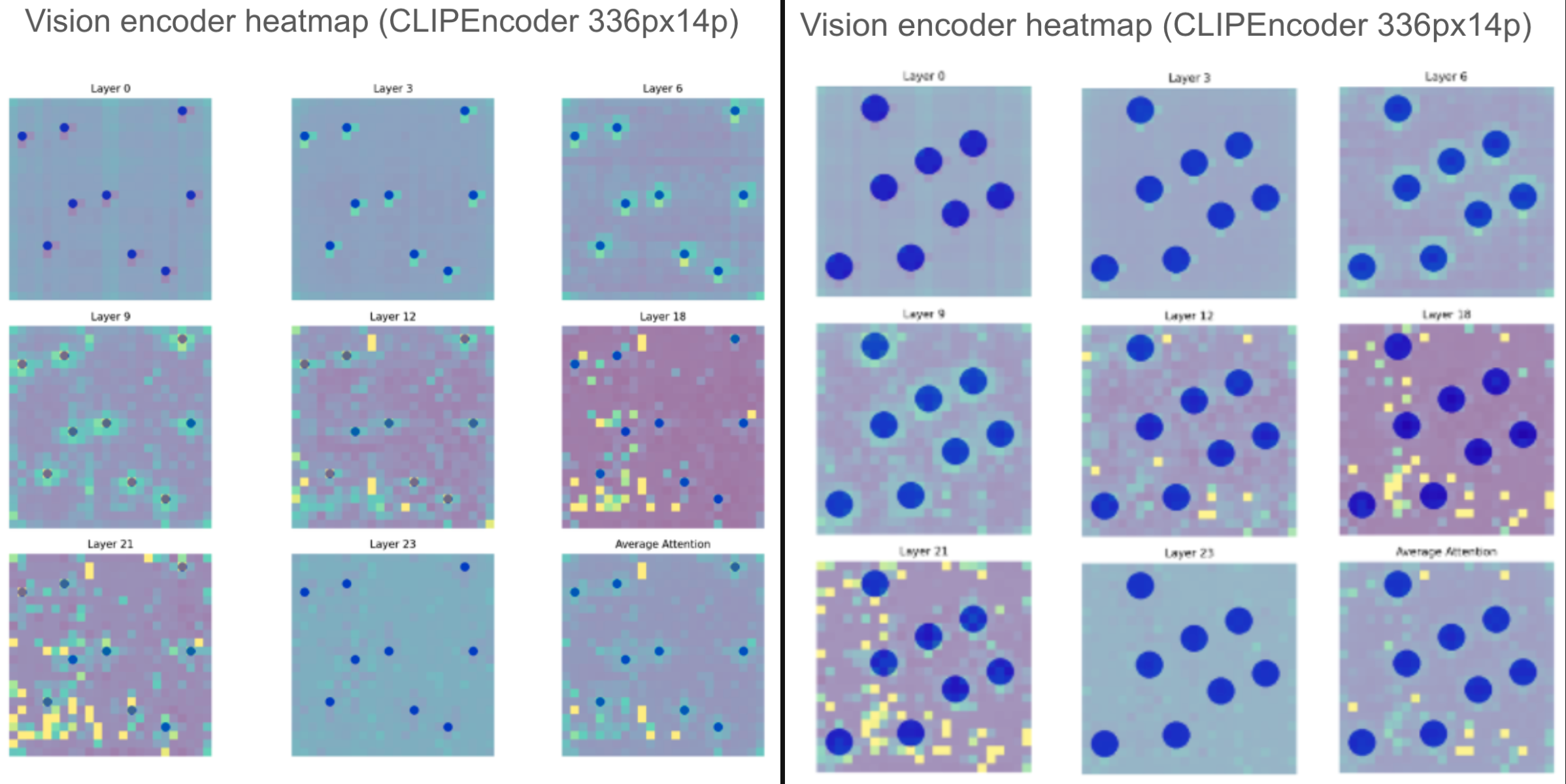}
    \caption{\textbf{Vision Encoder Heatmaps.} Activations in the early visual backbone are sharp and distinct, perfectly localizing the object instances. This confirms that the counting signal is present upstream, however, fading when entering the language space.}
    \label{fig:encoder_heatmaps}
    \end{figure}

\begin{enumerate}
    \item \textbf{Vision Encoder Fidelity.}
    To visualize the spatial signal, we first examine the attention maps extracted directly from the \textbf{Vision Encoder} (e.g., CLIP/SigLIP). As shown in Figure~\ref{fig:encoder_heatmaps}, these maps display sharp, high-frequency activations that are tightly clustered on the true object instances (blue dots). This visual evidence confirms that the backbone successfully ``sees'' and localizes the objects at the input stage.
    
    \item \textbf{Fused Token Diffusion.}
    In contrast, we analyze the attention maps from the \textbf{Fused Tokens} within the LLM (averaged over layers 15--25). We observe a dramatic loss of fidelity, the attention becomes diffuse and spatially unstructured, frequently spreading into background regions or aligning with uninformative tokens rather than semantically relevant content (i.e. the circles). This behavior suggests that cross-modal alignment further distorts the original representation of raw visual context, which explains the degradation that can be observed in Figure \textcolor{blue}{8} (Appendix).
\end{enumerate}

\textcolor{blue}{\textbf{Observation 5.}} \textit{Spatial detail exists upstream but washes out after fusion. The transition from sharp Encoder activations to diffuse Fused heatmaps confirms that the "visual attention sink" actively dilutes pixel-level evidence, replacing it with broad, non-specific attention distributions.}
\section{Methodology: Enforcing Visual Grounding via Modality Attention Share}
\label{sec:mas_method}

Our probing experiments in Sec.~\ref{sec:probing_analysis} revealed a critical failure mode: the "Visual Attention Sink." We observed that during reasoning, models disproportionately allocate attention to textual tokens, effectively ignoring the visual evidence. This insight drives our proposed intervention: \textbf{Modality Attention Share (MAS)}. We hypothesize that by explicitly constraining the model to "look" at the image during generation, we can bridge the Retention-Gap. This section formalizes MAS as a differentiable regularization objective that actively penalizes visual neglect.

\subsection{Quantifying Visual Reliance (MAS)}
\label{sec:mas_formulation}

To intervene, we first need a metric that is differentiable with respect to the model's weights. We define MAS as the ratio of attention mass allocated strictly to visual tokens. Let $\mathcal{V}$ be the set of visual token indices and $\mathcal{X}$ be the set of textual token indices. For a given layer $\ell$ and head $h$ at decoding step $t$, the attention distribution is $A^{(h,\ell)}_{t}$. The layer-wise visual share is:

\begin{equation}
\mathrm{MAS}_{\ell}
= \frac{1}{|\mathcal{T}|}\sum_{t\in\mathcal{T}}
\frac{\sum_{h}\sum_{j\in \mathcal{V}} A^{(h,\ell)}_{t\to j}}
     {\sum_{h}\sum_{j\in \mathcal{V}\cup\mathcal{X}} A^{(h,\ell)}_{t\to j}}.
\label{eq:mas_layer}
\end{equation}

Crucially, unlike post-hoc attention analysis, this formulation preserves the computational graph back to the query/key projections, allowing it to serve as a training signal.

\subsection{The Visual Constraint Loss}
\label{sec:mas_loss}

We frame visual grounding as a constrained optimization problem. We do not want to force the model to ignore text, but rather to ensure a \emph{minimum} level of visual consultation. We introduce a hinge loss that activates only when visual attention drops below a safety threshold $\tau$:

\begin{equation}
\mathcal{L}_{\text{mas}}
= \max\!\left(0,\ \tau - \mathrm{MAS}\right).
\label{eq:mas_penalty}
\end{equation}

This loss acts as an active guardrail:
\begin{itemize}
    \item If the model "hallucinates" (generates tokens with insufficient visual attention, $\mathrm{MAS} < \tau$), the loss spikes, generating gradients that push attention back to the image.
    \item If the model is sufficiently grounded ($\mathrm{MAS} \ge \tau$), the penalty vanishes, allowing standard cross-entropy to dominate.
\end{itemize}

The total objective combines standard instruction tuning with this grounding constraint: $\mathcal{L}_{\text{total}} = \mathcal{L}_{\text{CE}} + \lambda \mathcal{L}_{\text{mas}}$.

\subsection{Training Strategy and Data}
\label{sec:data_strategy}

To validate MAS, we construct a controlled instruction-tuning setup. We utilize the \textbf{FSC-147} dataset, converting its density maps into natural language conversations (e.g., "User: Count the cars." $\to$ "Assistant: I see 5 cars.").

\textbf{Targeting the Reasoning Phase.}
Not all tokens require visual grounding (e.g., stopwords like "The"). We therefore apply the MAS constraint selectively. Let the output sequence be $\mathbf{y} = (y_1, \dots, y_T)$. We define the target set $\mathcal{T}$ as the indices of the assistant's response only. This ensures we penalize "blind guessing" during the answer generation without disrupting the encoding of the user's prompt.

\subsection{Empirical Validation: MAS as an Intervention}
\label{sec:mas_ablation}

Our probing analysis indicates that counting failures often coincide with a \emph{visual attention sink}: during generation, attention mass drifts toward textual context, even when the answer is visually determined. MAS is a minimal intervention that targets this failure mode directly. Rather than changing architectures or introducing additional supervision, we add a hinge constraint that enforces a \emph{minimum} visual-attention share during answer generation (Sec.~\ref{sec:mas_loss}; $\tau{=}0.4$, $\lambda_{\text{mas}}{=}0.1$), and fine-tune for 10 epochs under the same instruction template.

\begin{table}[t]
\centering
\caption{\textbf{Impact of MAS Regularization.} Exact-match accuracy (\%). MAS improves validation accuracy for some backbones, but its effect is backbone- and dataset-dependent.}
\label{tab:mas_ablation}
\setlength{\tabcolsep}{4pt}
\resizebox{\linewidth}{!}{
\begin{tabular}{llccc}
\toprule
\textbf{Backbone} & \textbf{Training} & \textbf{Circles (Syn)} & \textbf{FSC-Val (Real)} & \textbf{FSC-Test (Real)} \\
\midrule
Ovis 2.5 & Pretrained & 53.8 & 6.3 & 3.9 \\
Ovis 2.5 & SFT & 84.9 & 17.5 & \textbf{16.6} \\
Ovis 2.5 & \textbf{SFT + MAS (Ours)} & \textbf{85.2} & \textbf{17.7} & 16.1 \\
\midrule
Qwen3-VL & Pretrained & \textbf{48.1} & 7.6 & 5.5 \\
Qwen3-VL & SFT & 18.2 & \textbf{20.3} & \textbf{14.0} \\
Qwen3-VL & \textbf{SFT + MAS} & 30.4 & 18.2 & 13.4 \\
\midrule
Intern3.5-VL & Pretrained & 44.6 & 0.0 & 6.5 \\
Intern3.5-VL & SFT & \textbf{63.1} & 16.9 & 14.4 \\
Intern3.5-VL & \textbf{SFT + MAS} & 53.7 & \textbf{17.7} & \textbf{14.9} \\
\bottomrule
\end{tabular}
}
\end{table}

Table~\ref{tab:mas_ablation} summarizes results across three backbones. For \textbf{Ovis-2.5}, MAS yields small but consistent improvements on in-distribution validation: \emph{Circles} increases from $84.9\%\rightarrow 85.2\%$ and \emph{FSC-Val} from $17.5\%\rightarrow 17.7\%$. However, performance on \emph{FSC-Test} decreases slightly ($16.6\%\rightarrow 16.1\%$), suggesting that enforcing higher visual attention share alone does not guarantee improved out-of-distribution generalization.

The effect is also \emph{backbone-dependent}. On \textbf{Qwen3-VL}, MAS improves the synthetic \emph{Circles} split compared to standard SFT ($18.2\%\rightarrow 30.4\%$), but reduces performance on \emph{FSC-Val/Test}. On \textbf{Intern3.5-VL}, MAS improves \emph{FSC-Val} ($16.9\%\rightarrow 17.7\%$) while slightly reducing \emph{Circles} and \emph{FSC-Test}. These mixed outcomes are informative: they indicate that (i) \emph{attention share} is a useful control knob, but (ii) it is not universally beneficial under a single fixed threshold, and it can trade off against other behaviors (e.g., output formatting or reliance on linguistic priors).

Overall, this ablation supports a conservative conclusion: MAS can act as a helpful inductive bias in some regimes by discouraging \emph{blind generation}, but attention regularization alone is not sufficient. In the remainder of the paper, we therefore treat MAS as evidence that the attention sink is \emph{mechanistically addressable}, while emphasizing that stronger interventions likely require grounding-aware step selection (e.g., targeting numeral tokens), normalization for token-length effects, and constraints that encourage \emph{where-to-look} alignment rather than only \emph{how-much-to-look}.



\section{Conclusion}
\label{sec:conclusion}

This work takes a diagnostic view of object counting in modern Vision--Language Models. Using \countingtricks{}, we show that strong language capabilities can create an \emph{illusion} of counting competence: models often produce plausible numbers, yet fail reliably under simple visual stressors and text-in-image conflicts. To move beyond answer-only evaluation, we probe the model stack and quantify where counting-relevant evidence is retained. Across open-weight models, we find that spatial signal is comparatively strong in early visual representations and projected tokens, but becomes substantially weaker at the LLM stage, where attention mass frequently drifts toward textual priors---a phenomenon consistent with a visual attention sink.

We further test whether this bottleneck is amenable to intervention. By introducing Modality Attention Share (MAS) as a hinge regularizer that enforces a minimum visual-attention budget during generation, we observe modest improvements on in-distribution validation for some backbones, alongside mixed effects on held-out test splits and across architectures. These results suggest two takeaways. First, ``blindness'' is not purely a limitation of the vision backbone: it can be shaped by how the LLM allocates computation during generation. Second, simply increasing attention to vision is not a complete solution; robust counting likely requires mechanisms that preserve spatial structure deeper into the reasoning layers and constraints that promote \emph{correct grounding} (where to look), not only \emph{more} grounding (how much to look).

We hope our benchmarks, probes, and analyses provide a practical foundation for studying grounding failures in VLMs, and for developing future architectures and training objectives that count from the visual world rather than from linguistic shortcuts.




{
    \small
    \bibliographystyle{ieeenat_fullname}
    \bibliography{main}
}


\newpage

\clearpage 
\appendix
\section{Appendix}
\label{appendix:main}

\subsection{Supplementary Experimental}
\label{appendix:detailed-baseline}

\noindent 
\begin{minipage}{\textwidth}
    \centering
    \begin{adjustbox}{width=\textwidth}
    \begin{tabular}{lrrrrrrrrrrrr}
    \toprule
    \textbf{Model} & 1A & 1B & 2A & 2B & 2C & 2D & 3A & 3B & 3C & 3D & 4A \\
    \midrule
    llava-1.5-7b & 16.9 & 16.5 & 16.4 & 15.3 & 16.7 & 16.0 & 17.1 & 15.2 & 16.7 & 14.9 & 17.8 \\
    llava-1.6-vicuna-7b & 18.0 & 20.0 & 18.5 & 18.3 & 19.4 & 20.2 & 18.9 & 19.4 & 17.9 & 19.7 & 20.5 \\
    internvl3-8b & 13.2 & 7.3 & 44.5 & 12.4 & 36.3 & 10.9 & 42.2 & 13.1 & 32.7 & 11.8 & 35.5 \\
    llama-3.2-11b & 34.0 & 24.3 & 42.0 & 30.1 & 39.0 & 27.7 & 34.7 & 22.6 & 33.8 & 24.7 & 57.5 \\
    gemma-3-4b & 39.9 & 33.8 & 40.1 & 36.4 & 35.5 & 33.2 & 39.4 & 35.9 & 36.0 & 31.7 & 36.0 \\
    qwen-2.5-VL-3b & 48.5 & 45.3 & 49.4 & 54.3 & 43.3 & 50.3 & 50.7 & 55.4 & 51.6 & 47.9 & 58.1 \\
    llava-onevision-qwen2-7b & 49.7 & 50.5 & 48.4 & 54.6 & 41.5 & 52.6 & 43.9 & 51.2 & 46.6 & 54.1 & 54.5 \\
    ovis-8b & 42.9 & 45.6 & 40.6 & 45.1 & 53.1 & 46.7 & 51.1 & 43.8 & 58.6 & 46.8 & 50.4 \\
    qwen-2.5-VL-7b & 56.3 & 59.9 & 62.8 & 65.2 & 59.0 & 63.5 & 52.4 & 56.7 & 55.3 & 59.9 & 63.0 \\
    phi4-multimodal & 71.4 & 51.6 & 71.1 & 56.2 & 69.3 & 55.7 & 71.0 & 60.7 & 68.4 & 59.1 & 61.6 \\
    \midrule
    \textbf{Average} & 39.08 & 35.48 & 43.38 & 38.79 & 41.31 & 37.68 & 42.14 & 37.40 & 41.76 & 37.06 & 45.49 \\
    \midrule
    \textbf{Model} & 4B & 4C & 4D & 5A & 6A & 7A & 8A & 9A & 9B & 10A & 10B \\
    \midrule
    llava-1.5-7b & 18.3 & 17.6 & 16.2 & 20.1 & 18.8 & 22.0 & 16.1 & 8.2 & 8.3 & 7.6 & 6.4 \\
    llava-1.6-vicuna-7b & 16.0 & 18.2 & 17.8 & 28.3 & 26.9 & 24.5 & 17.8 & 14.2 & 8.4 & 9.1 & 5.1 \\
    internvl3-8b & 14.5 & 32.3 & 14.8 & 65.5 & 60.7 & 52.3 & 45.0 & 19.1 & 8.9 & 17.1 & 8.9 \\
    llama-3.2-11b & 22.0 & 40.9 & 26.5 & 28.9 & 32.8 & 26.7 & 30.5 & 12.2 & 11.3 & 13.5 & 14.0 \\
    gemma-3-4b & 31.6 & 32.8 & 29.5 & 39.0 & 44.6 & 29.0 & 41.3 & 30.6 & 25.0 & 16.2 & 19.4 \\
    qwen-2.5-VL-3b & 51.4 & 50.3 & 55.3 & 77.2 & 78.9 & 72.0 & 83.2 & 21.2 & 8.2 & 14.6 & 22.1 \\
    llava-onevision-qwen2-7b & 52.8 & 50.1 & 56.8 & 66.7 & 73.9 & 65.5 & 75.4 & 42.4 & 15.0 & 41.7 & 23.4 \\
    ovis-8b & 38.7 & 60.6 & 49.0 & 73.2 & 82.2 & 75.0 & 80.6 & 65.4 & 32.7 & 41.8 & 33.8 \\
    qwen-2.5-VL-7b & 64.4 & 57.5 & 62.0 & 73.3 & 69.4 & 78.9 & 80.0 & 47.0 & 26.3 & 46.5 & 56.4 \\
    phi4-multimodal & 65.2 & 64.7 & 60.8 & 57.6 & 49.0 & 47.7 & 41.0 & 21.7 & 14.6 & 23.2 & 18.6 \\
    \midrule
    \textbf{Average} & 37.49 & 42.50 & 38.87 & 52.98 & 53.72 & 49.36 & 51.09 & 28.20 & 15.87 & 23.13 & 20.81 \\
    \midrule
    \textbf{Model} & 11A & 11B & 12A & 12B & 13A & 13B & 14A & 14B & 15A & 15B & \textbf{Avg $\uparrow$} \\
    \midrule
    llava-1.5-7b & 0.5 & 1.0 & 8.3 & 8.3 & 7.0 & 7.1 & 0.0 & 4.2 & 0.0 & 2.7 & \textbf{11.82} \\
    llava-1.6-vicuna-7b & 7.1 & 3.7 & 8.3 & 8.2 & 4.9 & 6.1 & 0.1 & 7.2 & 2.1 & 9.7 & \textbf{14.20} \\
    internvl3-8b & 9.6 & 6.0 & 17.9 & 9.6 & 16.5 & 8.5 & 9.9 & 4.1 & 9.4 & 4.6 & \textbf{21.72} \\
    llama-3.2-11b & 6.6 & 21.9 & 13.3 & 12.3 & 13.4 & 16.1 & 6.6 & 21.8 & 5.9 & 20.5 & \textbf{24.00} \\
    gemma-3-4b & 8.1 & 9.1 & 29.5 & 24.6 & 16.3 & 18.6 & 7.9 & 6.7 & 9.2 & 9.0 & \textbf{27.37} \\
    qwen-2.5-VL-3b & 3.8 & 0.3 & 13.1 & 7.4 & 16.7 & 14.1 & 6.5 & 0.3 & 0.4 & 0.5 & \textbf{36.01} \\
    llava-onevision-qwen2-7b & 41.8 & 21.2 & 43.2 & 17.2 & 40.3 & 24.9 & 40.1 & 20.1 & 39.7 & 22.2 & \textbf{44.44} \\
    ovis-8b & 42.4 & 25.5 & 66.4 & 34.3 & 40.3 & 33.2 & 38.4 & 26.4 & 41.9 & 26.2 & \textbf{47.90} \\
    qwen-2.5-VL-7b & 42.2 & 14.8 & 43.3 & 28.8 & 38.8 & 36.8 & 33.1 & 16.5 & 32.7 & 14.0 & \textbf{50.52} \\
    phi4-multimodal & 6.5 & 5.2 & 22.2 & 17.1 & 22.6 & 17.2 & 7.3 & 5.1 & 6.1 & 5.0 & \textbf{39.83} \\
    \midrule
    \textbf{Average} & 16.86 & 10.87 & 26.55 & 16.78 & 21.68 & 18.26 & 14.99 & 11.24 & 14.74 & 11.44 & \textbf{31.78} \\
    \bottomrule
    \end{tabular}
    \end{adjustbox}

    \vspace{10pt}
    \captionsetup{type=table, width=\textwidth, justification=justified}
    \caption{Complete accuracies over all 32 test cases, which case coding rule can be observed in the Sec. \ref{sec:bench}. Each code's evaluation or score is being represented by 1000 samples evenly distributed across the respective count range.}
    \label{tab:main_results_details}
\end{minipage}

\begin{figure*}[t]
    \centering
    \includegraphics[width=0.6\linewidth]{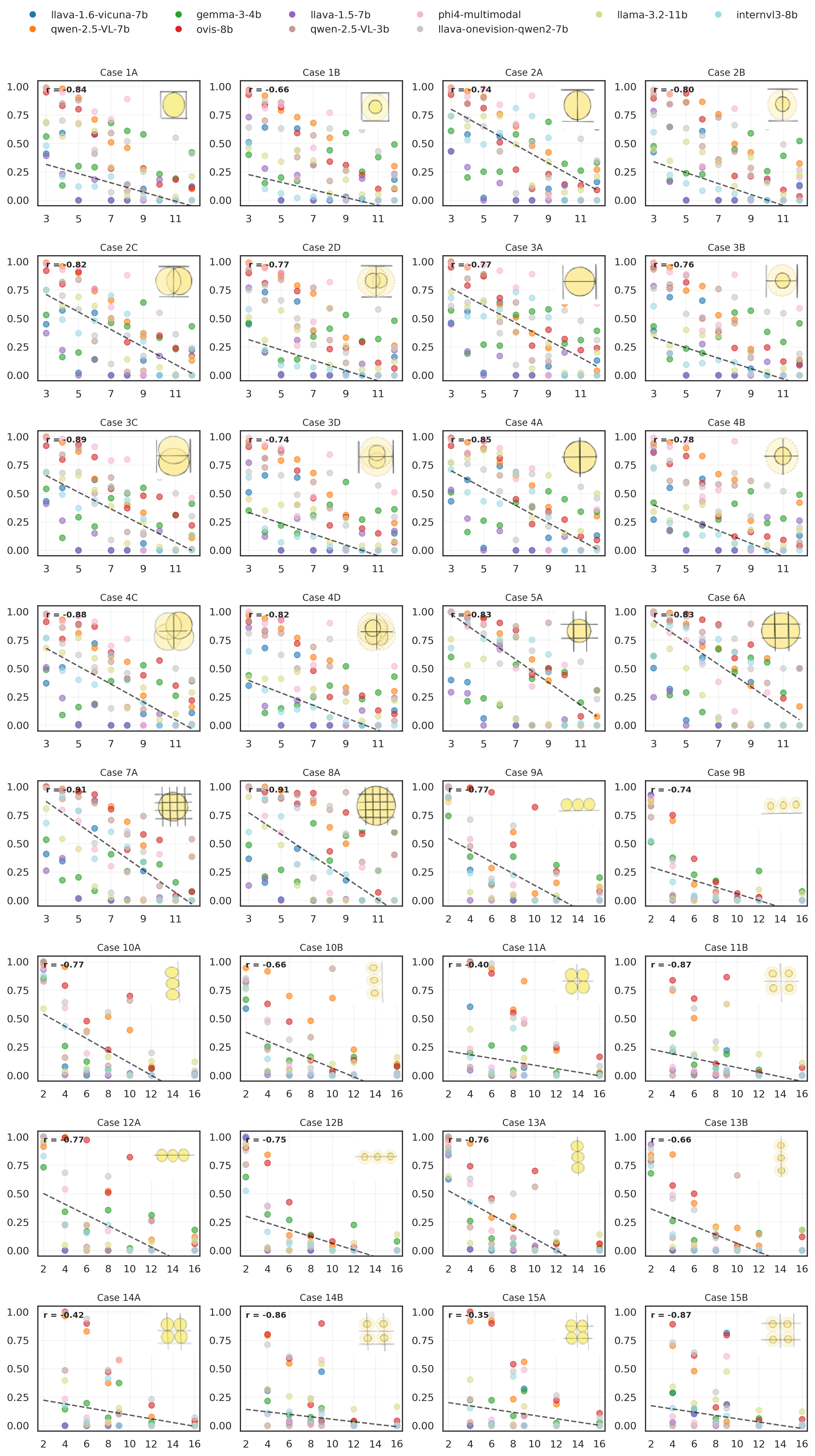}
    \caption{\textbf{Accuracy vs. Ground Truth Count.} \textit{Left:} Scatter plots show a consistent negative correlation ($r \approx -0.78$) between count magnitude and accuracy across diverse geometric cases. \textit{Right:} Detailed breakdown for LLaVA-1.5 and Qwen2.5-VL reveals specific ``blind spots'' (circled in red) where models achieve 0\% accuracy for specific numbers (e.g., 7, 11), evidencing strong linguistic priors.}
    \label{fig:acc_vs_count}
\end{figure*}

\begin{figure*}[t]
    \centering
    \includegraphics[width=0.95\linewidth]{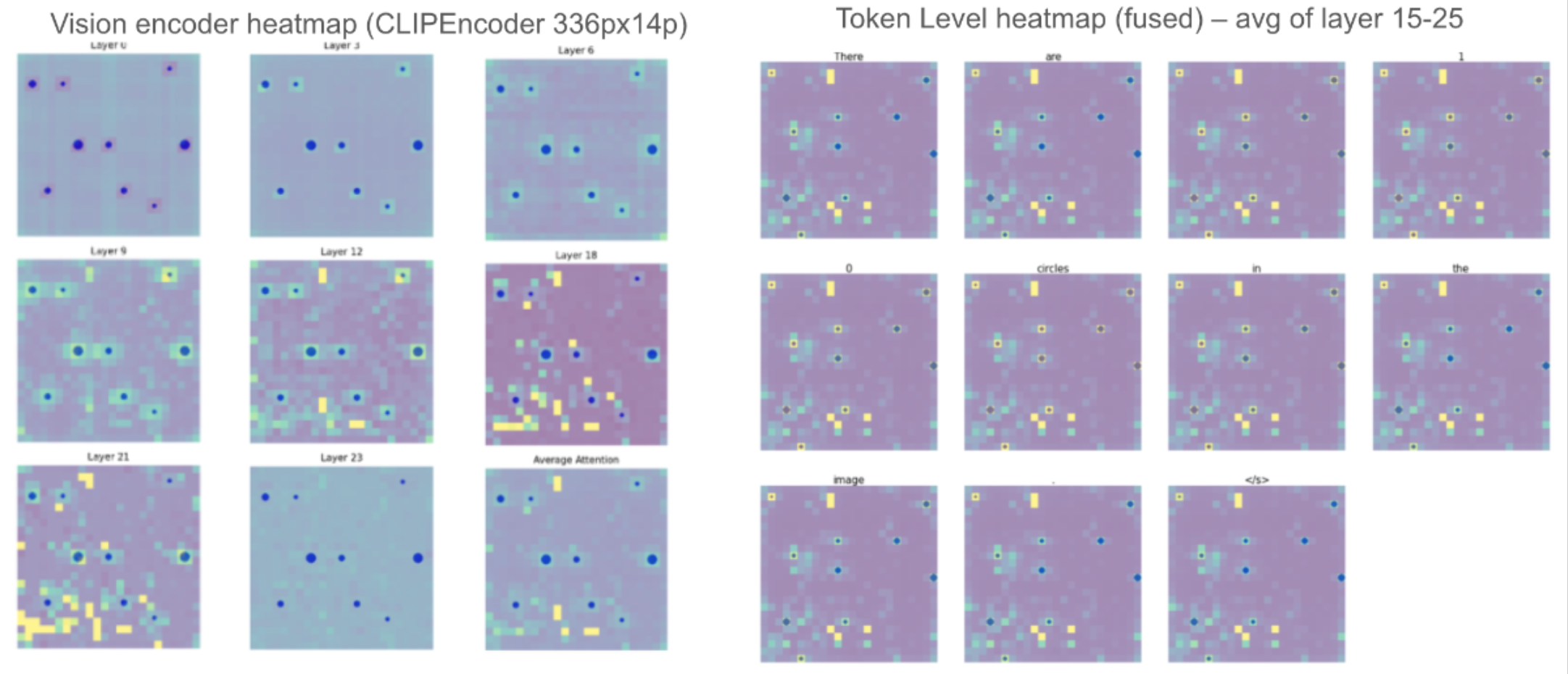}
    \caption{\textbf{Fused Token Heatmaps.} Averaged over deep LLM layers (15--25), the attention becomes diffuse and misaligned. The signal ``washes out,'' failing to retain the instance-level separation required for accurate counting.}
    \label{fig:fused_heatmaps}
\end{figure*}

\clearpage
\end{document}